\documentclass[]{templates/tADR2e}

\usepackage[hyphens]{url} 
\usepackage{soul} 
\usepackage{framed} 
\usepackage[normalem]{ulem} 
\usepackage{multirow}
\usepackage{mdframed}
\usepackage{enumitem}
\usepackage{booktabs}

\begin{document}


\jvol{38}
\jnum{17}
\jyear{2024}
\jmonth{September}

\title{
Reflectance Estimation for Proximity Sensing by Vision-Language Models:\\
Utilizing Distributional Semantics for Low-Level Cognition in Robotics
}

\author{
    Masashi Osada$^{a}$,
    Gustavo~A. Garcia Ricardez$^{a, \ast}$,
    Yosuke Suzuki$^{b}$, and
    Tadahiro Taniguchi$^{a}$\\
    \vspace{6pt}
    $^{a}${\em{College of Information Science and Engineering, Ritsumeikan University, Kusatsu, Japan}}\\
    $^{b}${\em{College of Science and Engineering, Kanazawa University, Kanazawa, Japan}}\\
    \vspace{6pt}
    \received{v2.1 released August 2024}
}

\maketitle


\begin{abstract}
    Large language models (LLMs) and vision-language models (VLMs) have been increasingly used in robotics for high-level cognition, but their use for low-level cognition, such as interpreting sensor information, remains underexplored. In robotic grasping, estimating the reflectance of objects is crucial for successful grasping, as it significantly impacts the distance measured by proximity sensors. We investigate whether LLMs can estimate reflectance from object names alone, leveraging the embedded human knowledge in distributional semantics, and if the latent structure of language in VLMs positively affects image-based reflectance estimation. In this paper, we verify that 1) LLMs such as GPT-3.5 and GPT-4 can estimate an object's reflectance using only text as input; and 2) VLMs such as CLIP can increase their generalization capabilities in reflectance estimation from images. Our experiments show that GPT-4 can estimate an object's reflectance using only text input with a mean error of 14.7\%, lower than the image-only ResNet. Moreover, CLIP achieved the lowest mean error of 11.8\%, while GPT-3.5 obtained a competitive 19.9\% compared to ResNet's 17.8\%. These results suggest that the distributional semantics in LLMs and VLMs increases their generalization capabilities, and the knowledge acquired by VLMs benefits from the latent structure of language.
    
    \medskip
\end{abstract}


\section{Introduction}
\label{sec:introduction}

In recent years, large language models (LLMs) and vision-language models (VLMs) have been used in robotics for their high-level cognitive capabilities such as task planning and language understanding~\cite{pmlr-v164-shridhar22a, ahn2022can, khandelwal2022simple, melnik2024exploring}.
However, the usage of their low-level cognitive capabilities for real-world applications in the robotics field, such as sensor information interpretation, has not been thoroughly explored\footnote{We can find a few examples such as~\cite{loyola2023perceptual,kawakita2023comparing,rubungo2023llmprop}.}.
We argue that such low-level cognitive capabilities can be used in robotic grasping.

For a robot to grasp an object with various shapes and properties, it is possible to use proximity sensors to adjust the end effector pose before contact.
However, as objects reflect the light emitted by the sensors according to their shape and properties, the sensors cannot accurately measure the distance unless such \textit{reflectance} is known.
Therefore, estimating the object's reflectance is crucial for successful grasping and this requires interpreting the sensor information.

When attempting to estimate an object's reflectance, conventional deep learning-based supervised learning systems are trained with labeled data for object categories and estimated values of their reflectance as teaching signals~\cite{garcia2023}.
In such systems, the only clue for generalizing the learning results is the similarity of input images. 
Therefore, it is not possible to use the linguistic semantics of category names themselves for generalization.

In contrast, LLMs and VLMs implicitly form knowledge from the corpus based on distributional semantics~\cite{Harris1954DistributionalS} (i.e., how language preserves the relationship between concepts represented by words in a sequence), which could potentially be used for generalization.
With LLMs and VLMs, it is possible to generalize knowledge about various vocabularies from the latent semantic relationships inherent in the language corpus~\cite{brown2020_fewshot, dall-e}. 
This generalization is not limited to the finite number of labels, which is often given in the training dataset of supervised learning-based estimation systems.
Therefore, we focus on how LLMs and VLMs can leverage distributional semantics to interpret sensor information for low-level cognition in robotics.

More specifically, we aim to answer the following research questions:
\begin{enumerate}[label=\textbf{RQ\arabic*}:, leftmargin=1.5cm]
    \item \textit{Can LLMs estimate reflectance for proximity sensing just from the object name by using distributional semantics without any image features?}
    \item \textit{Can the distributional semantics embedded in VLMs improve the performance of image-based reflectance estimation for proximity sensing with deep neural networks?}
\end{enumerate}

In this paper, we demonstrate that even when estimating object-dependent reflectance, learning from the language corpus through the knowledge inherent in distributional semantics is useful, clarifying the advantages of utilizing distributional semantics in low-level cognition in robotics.
We experimentally investigate the ability of LLMs and VLMs trained with only text and with text-image pairs to estimate the reflectance of objects and shed light on the capabilities of these models to exhibit low-level cognition.
Furthermore, we use a method based on Contrastive Language-Image Pre-Training (CLIP)~\cite{clip}, a type of VLM that has been contrast-trained on a dataset consisting of an image and a string describing that image, and task it with reflectance estimation.
Next, we compare the accuracy of the estimated reflectance by LLMs and VLMs, and the conventional approach.
Finally, we experimentally verify the effectiveness of the proposed method for robot grasping using a two-finger gripper.

This paper presents a case where a high processing speed is not necessary by following the commonly used sensing-perception-planning paradigm, as opposed to servoing-like applications where such speed is paramount.
Once the visual or textual input is given, the corresponding method estimates the reflectance once for the target object, which serves as a calibration for the proximity sensor.
Then, a separate high-frequency sensing-control cycle could be carried out to move the end effector and grasp the target object.

The main contributions of this paper are the following:
\begin{itemize}
    \item We verify that LLMs and VLMs can be used for reflectance estimation, which suggests that objects and their reflectance are implicitly related in language due to distributional semantics.
    \item We verify that LLMs can estimate reflectance at a competitive level compared to conventional vision-based recognition methods without additional training using image data.
\end{itemize}

The rest of the paper is organized as follows.
Section~\ref{sec:background} introduces the related works.
Section~\ref{sec:proposed_solution} details the evaluated methods.
Section~\ref{sec:experiments} presents the experimental results and offers a discussion.
Finally, Section~\ref{sec:conclusion} concludes this paper.

\begin{figure}[t]
  \centering
    \includegraphics[width=\textwidth]{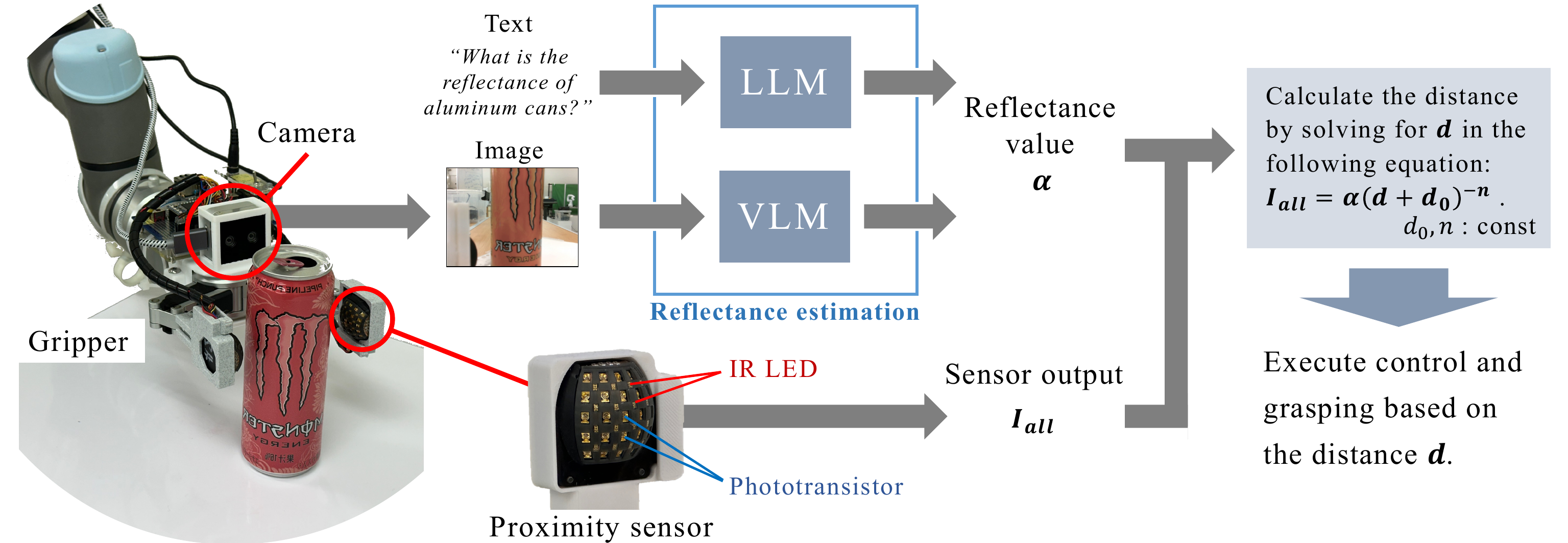}
    \caption{Reflectance estimation for proximity sensing. The LLMs and VLMs estimate the reflectance of objects whether from text (LLMs) or from images (VLMs). Together with the sensors' output and the intrinsic parameters of the sensors, the estimated reflectance is used to measure the distance to target objects in preparation for grasping.}
    \label{fig:system}
\end{figure}


\section{Background}
\label{sec:background}

First, we introduce previous works on using sensing information before contact in the context of robotic grasping.
Here, we show the advantages of the type of sensors of our focus (i.e., proximity sensors), as well as relevant works demonstrating them.
Second, we introduce the need for the interpretation of sensing information and argue that the generalization capabilities of LLMs and VLMs derived from distributional semantics can be used for such purpose.

\subsection{Robotic Grasping with Proximity Sensors}

On the one hand, for a robot to handle objects according to their various characteristics, it is desirable that the grasp pose is adjusted before contact.
Proximity sensors can be used for this purpose~\cite{navarro2022}.
These sensors that use optical reflection are compact and have a fast response time, which makes them suitable for mounting on fingertips and for providing feedback for motion control.
Koyama et al.~\cite{koyama2015} attached a sensor to the fingertip of the gripper to constantly measure the position of the gripper relative to the object, thereby enabling the robot to grasp objects according to their shape.
The sensor attached to the fingertip of the gripper is a sensor that emits infrared light from the fingertip and receives the reflected light from the object, as shown in Fig.~\ref{fig:system}.
This proximity sensor emits infrared light from the fingertip and calculates the distance based on the intensity of the infrared light reflected from the object.

In the proximity sensor used by Koyama et al., when the infrared light emitted by the infrared LED is reflected from an object, the phototransistor observes a reduced amount of infrared light according to the distance between the object and the sensor and the properties of the object's surface~\cite{koyama2015}.
The phototransistor takes light as input and returns an electric current, and the degree of the current value depends on the amount of input light.
Koyama et al. modeled the relationship between the obtained current value and distance as $I_{\text{all}} = \alpha (d + d_{0}) ^ {-n}$,
where $I_{\text{all}}$ is the current value obtained by combining the outputs of all phototransistors, $\alpha$ is the reflectance of infrared light of each object, and $d$ is the distance between the object and the sensor. 
In addition, $d_0$ and $n$ are intrinsic parameters.
By attaching a sensor to the fingertip of the gripper and constantly measuring its relative position to the object, they developed a method to control the movement of the fingertip for grasping according to the shape of the object and to carefully move the fingertip when the object and the fingertip are close to each other.
Their work proposed a generic approach to grasping for object shape and softness.

Hirai et al. proposed a method that complements the image input from the camera with proximity information for areas that cannot be seen by the camera, and showed that object grasping can be performed faster and with a higher success rate~\cite{hirai2018}.
In addition, Suzuki et al. proposed a method to constantly evaluate the stability of grasping by accurately acquiring the shape of the object using the same sensor and showed that the success rate of grasping is improved for a variety of objects~\cite{suzuki2022}.
Hsiao et al. introduced proximity sensors integrated into the fingertips of a BarrettHand which, combined with a probabilistic model and a hierarchical reactive controller, improved grasping dynamically~\cite{hsiao2009}.
They utilized proximity sensors to make real-time adjustments to grasp points without the need for premature object contact or regrasping strategies.

The relationship between the intensity of the infrared light measured by the proximity sensor and the actual distance depends on the physical properties of the surface of each object and thus requires \textit{a priori} information on the degree of reflection of the infrared light from each object.
Konstantinova et al. proposed a calibration method using RGB information obtained from a camera~\cite{konstantinova2016}.
The accuracy of the proximity sensor is enhanced using the RGB information corresponding to the object to be grasped and then this data is used to obtain the correct calibration for the proximity sensor.
Hasegawa et al. proposed a method to estimate object-dependent parameters when used in conjunction with a reflectivity-independent Time-of-Flight sensor~\cite{hasegawa2020}.
Their study fuses close-range and long-range sensors to overcome the limitations of each type of sensor.

From the model of Koyama et al., the equation of $I_{\text{all}}$ indicates that the process of calculating the distance using the proximity sensor depends on the reflectance of the infrared light of the object of interest~\cite{koyama2015}.
The calculation of the reflectance of each object requires that the distance between the sensor and the object is known, and it is difficult to accurately determine the reflectance for unknown objects.
The method of Hirai et al.~\cite{hirai2018} uses manual input of the infrared reflectance of the target object, but this method is labor-intensive and may lack generalization performance for unknown objects.

To address this issue, Garcia et al. developed a system that predicts the infrared reflectance of a target object from image input using a CNN~\cite{garcia2023}.
Specifically, the system uses representative CNNs, VGG16~\cite{vgg} and ResNet101~\cite{resnet}, pre-trained on ImageNet, a large image dataset, and then performs fine-tuning to adapt the system to the task at hand.
In the inference process, they assume that the reflectance of infrared light follows the category information of each object, and then calculate the expected reflectance of the object by summing the inferred likelihoods for each category with the weighted reflectance of the infrared light corresponding to each category.
This means that the classification task is solved when the reflectance of each object calculated from the actual distance measured using the proximity sensor is considered as an independent category.
This allows for efficient learning by taking into account the material of the object. 
While it is possible to learn efficiently considering the material of the object, it may be highly dependent on the data to be trained, and generalization performance for objects of unknown material may not be guaranteed.

\subsection{Language Models for Low-level Cognition}

On the other hand, even though LLMs have been attracting considerable attention in various fields, there is no satisfactory explanation for their extensive knowledge about our world and their ability to act appropriately, according to Mahowald et al.~\cite{mahowald2024dissociating}.
In general, the capabilities of LLMs are often discussed from a computational perspective, focusing on the transformer network structure~\cite{NIPS2017_3f5ee243}.
However, it is known that language preserves the relationship between concepts represented by each word in a sequence, which is known as distributional semantics~\cite{Harris1954DistributionalS}.
Distributional semantics studies the quantification and categorization of semantic similarities between linguistic items based on their distribution in language data~\cite{lenci2018}.

The knowledge embedded in LLMs arises from distributional semantics, an intrinsic part of the language formed by human society, as explained from the perspective of collective predictive coding in symbol emergence in~\cite{Taniguchi2024CollectivePC}.
VLMs extend the principles of distributional semantics to multimodal contexts, where language is combined with visual information.
By integrating distributional semantics with visual representations, VLMs can learn rich, generalized semantic knowledge that contains both linguistic and visual cues.

Gurnee and Tegmark demonstrated that LLMs learn representations of space and time across multiple scales~\cite{gurnee2024language}. 
Marjieh et al. investigated the extent to which LLMs can predict human sensory judgments~\cite{marjieh2023large}. 
The study reveals significant correlations between the model-generated judgments and human data across different domains, such as color and pitch, indicating the potential for extracting perceptual information from language. 

Loyola et al. showed that there is significant correspondence between the human perceptual color space and the feature space found by language models~\cite{loyola2023perceptual}.
Kawakita et al. compared the color similarity structures of humans (both color-neurotypical and color-atypical participants) with two GPT models, GPT-3.5 and GPT-4~\cite{kawakita2023comparing}. 
The results indicated that the similarity structures of color-neurotypical humans align well with those of GPT-4, and to a lesser extent with GPT-3.5.

To predict the physical and electronic properties of crystalline solids from their text descriptions, LLM-Prop has been proposed~\cite{rubungo2023llmprop}.
Their method outperformed existing methods in predicting properties like band gap, unit cell volume, and more.
This shows that LLMs are capable of predicting various physical properties of objects.

Tang et al. proposed to use the semantic knowledge from LLMs to realize zero-shot generalization of novel concepts to perform task-oriented grasping~\cite{graspgpt}.
This work seems to partially use LLMs for low-level cognition, as they generalize grasping from known object classes and known tasks by using the semantic knowledge of LLMs.
Gao et al. proposed a method to physically ground a VLM by fine-tuning it with human annotations~\cite{pgvlm2024}.
This work partially uses VLMs for low-level cognition since physical properties such as mass, fragility, and deformability are annotated to fine-tune the model for more accurate estimation, which is then used for high-level cognition.


\section{Methods}
\label{sec:proposed_solution}

\begin{figure}[t]
  \centering
    \includegraphics[width=\textwidth]{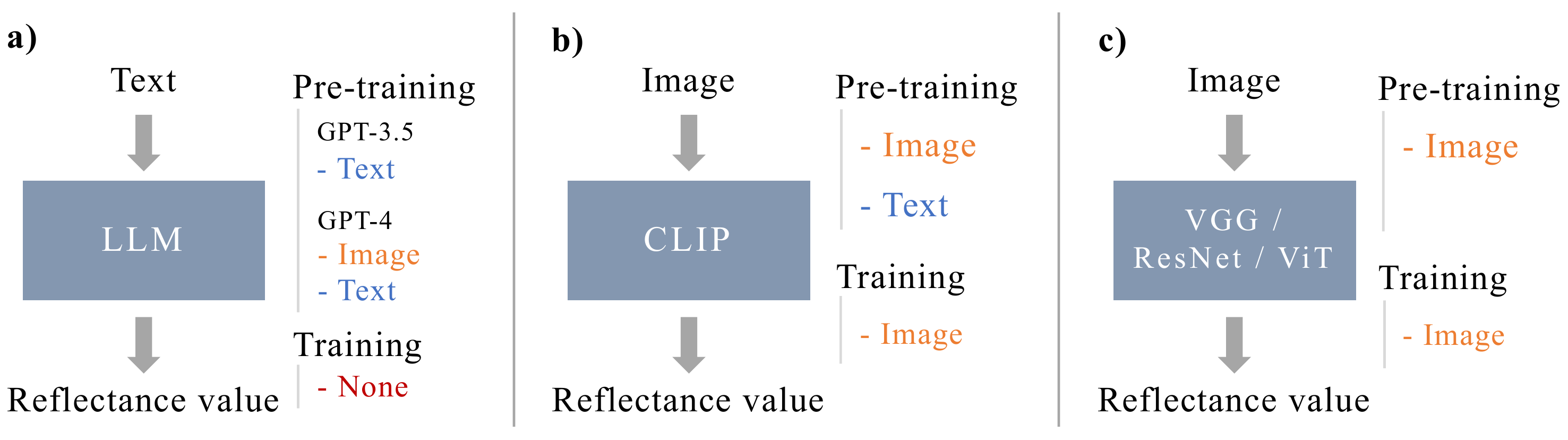}
    \caption{Overview of the three groups of methods used in the experiments.
    a) LLMs undergo pre-training that is either text-based or involves both text and images, without additional training.
    b) VLMs employ text and image pre-training, with additional image-based training. 
    c) Compared methods use image-only pre-training and additional image training.}
    \label{fig:model}
\end{figure}

This section introduces the compared methods used for the reflectance estimation task.
GPT-3.5 is a model trained only on text, while GPT-4 and CLIP are trained with images and text.
As GPT-4 and CLIP are trained through different mechanisms, the former is considered an LLM, while the latter is considered a VLM.
It is worth mentioning that even though GPT-4 is known to have been trained with images too, its input for inference remains to be only text.
To clarify the effect of distributional semantics, we also compared the aforementioned methods to the image-only trained methods VGG16, ResNet101 and ViT-B/32~\cite{vit}.
Fig.~\ref{fig:model} shows a summary of the compared methods.

\subsection{GPT-3.5 and GPT-4}

We use ChatGPT\footnote{We use the versions gpt-3.5-turbo-0125 and gpt-4-0125-preview.}, a chat service based on GPT-3.5 and GPT-4, as the LLMs.
First, since the input to ChatGPT is a string, we manually assigned a string of one or two words that describe the images in the dataset. 

The methodologies employed in this study are constructed through the application of Few-Shot Prompting and Chain-of-Thought techniques.
Few-Shot Prompting refers to the method of providing a prompt with several example answers~\cite{brown2020_fewshot}.
This approach, compared to Zero-Shot Prompting~\cite{ling2017_zeroshot} where no examples are provided, offers stronger guidance through the prompt and has been reported to enhance the accuracy of outputs from LLMs across a variety of inference problems~\cite{min2022_fewshot,cobbe2021_chainofthought_metric}.
Chain-of-Thought is a technique that involves explicitly or implicitly indicating the sequence of reasoning within the prompt, enabling more advanced inference than Few-Shot Prompting~\cite{wei2023_chainofthought}.

In this paper, we initially apply the concept of Few-Shot Prompting by providing the prompt with pairs consisting of a descriptive string for each image in the training dataset and the corresponding reflectance. 
Subsequently, based on the Chain-of-Thought approach, the method of thinking leading up to the answer is explicitly included in the prompt. 
First, the characteristics of light reflection on the object's surface are inferred based on the object's name. 
Next, by comparing the surface features and the sample reflectance provided in Few-Shot, a possible range of infrared reflectance is estimated. 
Finally, the most plausible value of infrared reflectance is provided as the answer. 
This method expects to obtain answers considering the material of the object. 
Additionally, this method can also predict the surface material implied by the object's name; for example, from the name `Coffee,' it is experimentally confirmed that the surface is treated as equivalent to plastic if it is presumed to be in a plastic bottle, and the reflectance is answered accordingly.
This leverages the inherent property of language models where the output of the LLM is influenced by the preceding text.

By doing this, we encourage the prediction of the reflectance values based on the material and reflective characteristics of each object.
Finally, only the descriptive strings explaining each image in the test dataset are provided.
Since Few-Shot Prompting initially indicates that the infrared reflectance follows the string representing the object, ChatGPT consequently outputs the reflectance for each string in the test dataset.
A summarized example of the prompts used in this work is shown in Appendix~\ref{sec:appendix-prompt} and in our repository\footnote{https://github.com/Osada-M/ReflectanceEstimationByChatGPT}.

\subsection{CLIP}

The purpose of using CLIP in our research is based on the expectation that linguistic information retains details about light reflection, allowing for more accurate predictions than when using images alone.
In other words, we expect the latent structure of language to affect the knowledge acquired by CLIP, as postulated by distributional semantics, leading to higher accuracy in the reflectance estimation from images.

CLIP~\cite{clip} is a deep learning model that consists of two encoders: an image encoder and a text encoder.
In the training process of CLIP, the visual information and linguistic information are contrasted using a dataset consisting of images and text strings describing the images, which are collected from a large number of images on the Web.
Specifically, the image encoder and text encoder compute the cosine similarity of the extracted feature representations, and then proceed to increase the similarity if the input image/text pair is correct, and decrease the similarity if the pair is different.
This makes it possible to extract common feature representations from images and text, respectively, so it is possible to consider the extraction of linguistic information from images or visual information from language.

The deep learning model used in the current work consists of a CLIP image encoder and two fully-connected layers.
The number of units of all connected layers is set to 32 and 1, respectively. 
The second fully-connected layer has only one unit, so its function is to convert the output of the first layer (a 33-dimensional vector including bias) into a single number (a scalar).
The learning process is performed in such a way that this scalar value corresponds to the desired reflectance.
In general, when a pre-trained CNN is applied to individual tasks using a dataset such as ImageNet for category prediction, it is expected that the category prediction information obtained through pre-training will be useful for the desired inference.
On the other hand, when CLIP is applied to individual tasks, it is expected that the target task can be represented in a more complex way because common feature representations are extracted from images and language. 

Although there are various versions of CLIP depending on the model size and patch size, we use two versions: ResNet101~\cite{resnet} and Vision Transformer (ViT-B/32)~\cite{vit}.
However, as a pre-training method for CLIP, we use the weights that have already been trained on 400 million pairs of images and text strings collected from the Web, as used in the original paper~\cite{clip} by Alec et al., where they refer to it as WebImageText.
We use the official implementation indicated in the same paper\footnote{https://github.com/openai/CLIP}, with the pre-trained weights of each model\footnote{ViT-B/32: openai/clip-vit-base-patch32, ResNet101: models/clip-resnet-101-visual-float16}.
Furthermore, while CLIP consists of an image encoder and a text encoder, in this paper, we consider only the input of images and utilize the output from the image encoder exclusively.
As CLIP acquires representations of features common to both language and images through Contrastive Learning, there is linguistic information implicitly related to the input image.
In other words, the output of the image encoder for an image is pre-trained to be similar to the output of the text encoder for the text describing the same image, meaning that the output of the image encoder represents complex features with language as contextual information.


\section{Experiments}
\label{sec:experiments}

We conduct experiments to estimate the reflectance of objects using GPT-3.5, GPT-4 and CLIP with ResNet101 and ViT-B/32.
We evaluate the estimation accuracy by comparing the aforementioned methods to conventional image-only VGG16, ResNet101 and ViT-B/32.
We set the loss function as the mean squared error, which is widely used in regression tasks.

\begin{figure}[t]
  \centering
    \includegraphics[width=160mm]{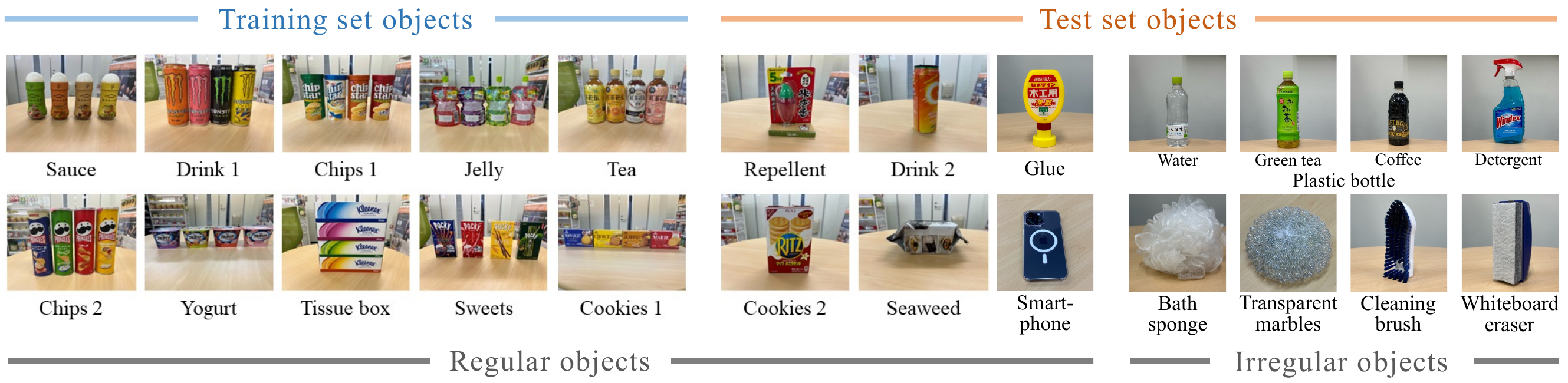}
    \caption{Objects used in the experiments. 
    The objects on the left were used during the training phase (training set), while those on the right were used during the testing phase (test set). 
    The latter are unknown/unseen to the compared methods.
    Besides regular objects, we also conducted experiments on irregular objects.
    }
    \label{fig:objects}
\end{figure}

\begin{figure}[t]
  \centering
    \includegraphics[width=75mm]{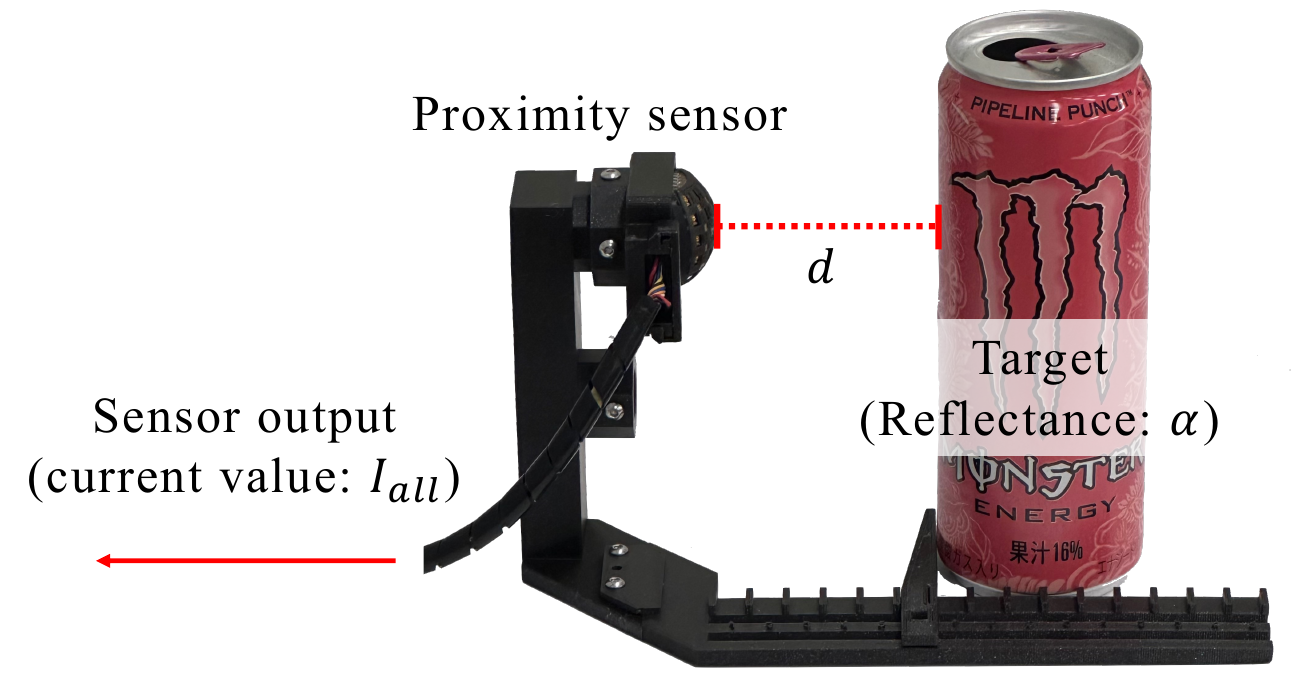}
    \caption{Setup to obtain the reflectance (ground truth) of objects at known distances.
    Given $I_{\text{all}} = \alpha (d + d_{0}) ^ {-n}$, where $I_{\text{all}}$ is the current value obtained by combining the outputs of all phototransistor (measured) and $d_0$ and $n$ are intrinsic parameters (known), we can calculate the reflectance $\alpha$ if the distance between the object and the sensor $d$ is also known.
    We use the least squares method to obtain the ground truth value of $\alpha$ by placing the object at various distances.
    }
    \label{fig:measuring-reflectance-setup}
\end{figure}

\subsection{Dataset}
\label{sec:dataset}

The dataset used in this paper consists of 324 pairs of images of 54 different objects taken from 6 different distances and the reflectance values of infrared light for each object.
From these objects, 40 are treated as training data (\textit{known objects}) and 14 as test data (\textit{unseen objects}).
In other words, the unseen objects are unknown to the evaluated methods.
Fig.~\ref{fig:objects} shows the objects in the training set and in the test set.

Besides conventional or regular objects, we also conduct experiments on irregular objects.
This is because irregular objects are particularly challenging for proximity sensors because their non-uniform surface's features and/or their transparency cause atypical readings from the sensors.

The true value of the reflectance of infrared light was calculated using a proximity sensor, and $I_{all}$ was calculated from the relationship between the current output from the sensor when the sensor was moved between 5 and 30~mm at regular intervals from the object and the actual distance.
The setup to collect such data is shown in Fig.~\ref{fig:measuring-reflectance-setup}, while examples of the proximity sensor output are shown in Appendix~\ref{sec:appendix-sensor-outputs}.

\subsection{Settings for Training}
\label{sec:parameters}

In this experiment, we used several deep learning models to make predictions based on image inputs. Specifically, we utilized ResNet101 and ViT-B/32 as backbones, and compared cases where only images were used for pre-training with cases where CLIP, which includes both images and language, was used.
Additional training is conducted to adapt these models to our task.

First, we attach a fully connected layer after the final intermediate layer of these pre-trained models. 
This is to align the model's predictions with the reflectance format. 
Since the training set is small, as transfer learning, we freeze the weights of the backbone and only update the weights of the added fully connected layer. 
This is to avoid destroying the weights obtained from pre-training with the small amount of data.

During training, we update the weights using Adam~\cite{adam} with a learning rate $1\times10^{-3}$. 
Also, since the fully connected layer is not pre-trained, we make significant weight changes at the start of training using cosine-annealing~\cite{cosine_annealing}. 
The number of epochs is 200.

\subsection{Reflectance Estimation}
\label{sec:reflectance-results}

We evaluate the accuracy of reflectance estimation by inputting images to VGG16, ResNet101, ViT-B/32, CLIP (ResNet101) and CLIP (ViT-B/32), and by inputting only text to GPT-3.5 and GPT-4.
With each method, we perform 6 trials per object.
We perform the experiments with 40 known objects and 14 unseen objects; the latter are not used for training.

\begin{figure*}
     \centering
     \includegraphics[width=\textwidth]{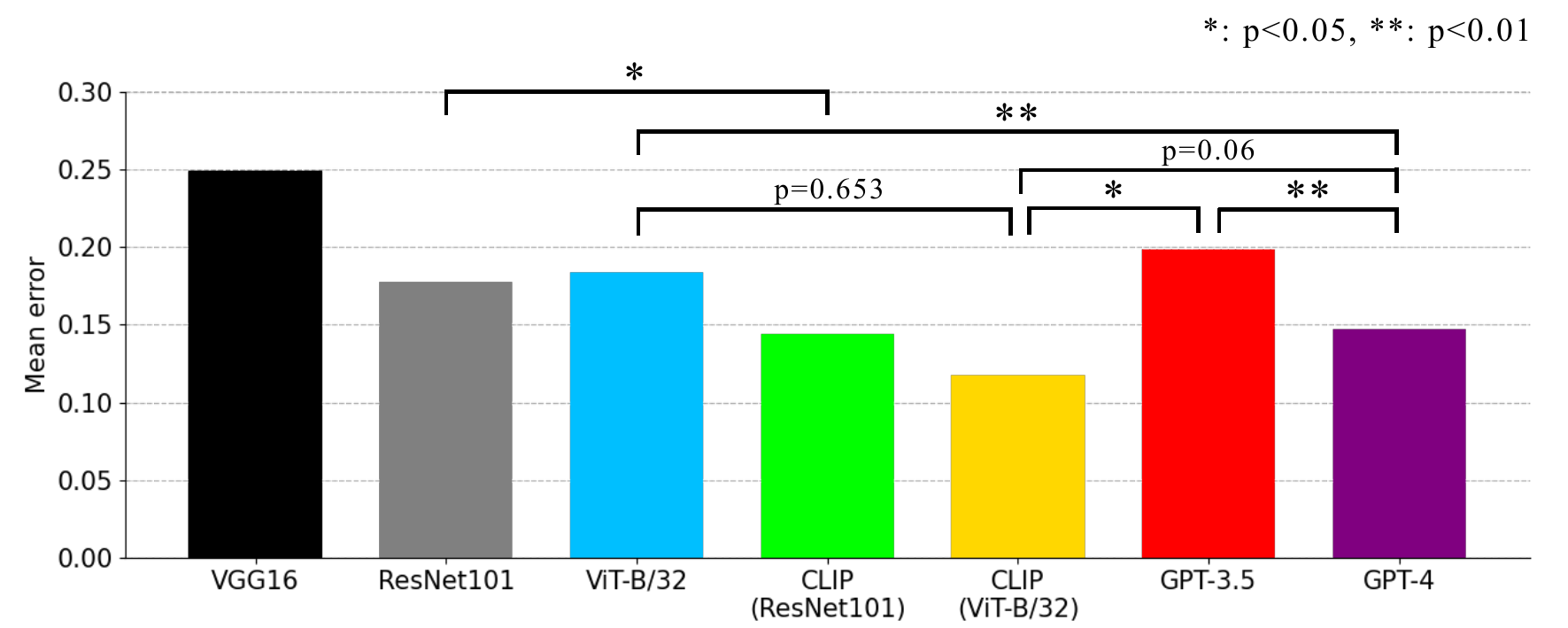}
     \caption{Mean error of the compared methods.} 
     \label{fig:significance}
\end{figure*}

\begin{figure*}
     \centering
     \includegraphics[width=\textwidth]{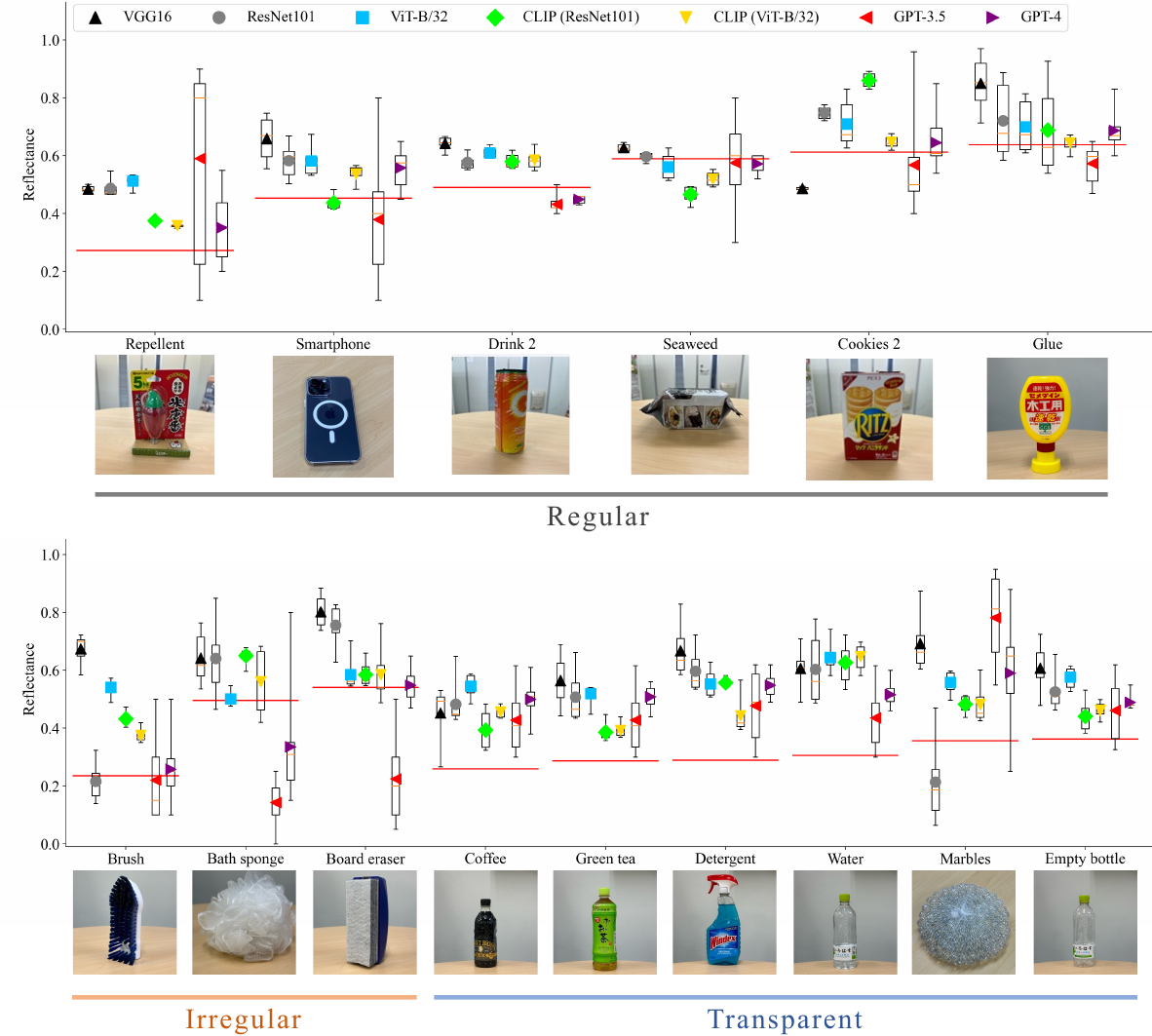}
     \caption{Estimated reflectance for unseen objects (test set).}
     \label{fig:visual_compare}
\end{figure*}

Table~\ref{table:mae} shows the mean error between the ground truth and the predicted value of each method, while Fig.~\ref{fig:significance} shows also the statistical significance between the compared methods.
Fig.~\ref{fig:visual_compare} shows the results of reflectance estimation for the unseen objects (test set).
It is worth mentioning that in the case of known objects (training set), as ChatGPT was given the true values directly in the prompts, the predictions for the training data returned exactly the true values (see Appendix~\ref{sec:appendix-reflectance-known}).
Therefore, we focus only on evaluating the accuracy of unseen objects (test set).

Table~\ref{table:mae} shows that the prediction using GPT-4 has higher accuracy compared to conventional methods, namely, VGG16, ResNet101 and ViT-B/32, which use solely images.
When comparing GPT-3.5 and GPT-4, GPT-4 has higher accuracy.
GPT-3.5 has higher accuracy than VGG16 and similar though inferior accuracy than ResNet101 and ViT-B/32.
Although the structure and specific training processes of both models have not been disclosed at this time, the Technical Report on GPT-4~\cite{gpt-4} shows that the number of parameters and training data for GPT-4 has increased significantly compared to GPT-3.5, and that GPT-4 is trained not only with language inputs but also with images.

The results show that the CLIP-based methods have a relatively high prediction accuracy for unseen objects compared to conventional methods.
This is because the models can extract common information from images and language, allowing for more complex feature representations in the reflectance prediction model as a whole.

Fig.~\ref{fig:visual_compare} shows that with the exception of the \textit{Repellent}, GPT methods achieve a relatively high accuracy.
For the set of irregular objects, GPT-3.5 tends to underestimate the reflectance more than GPT-4.
For transparent objects, most methods overestimate the reflectance, with only ResNet101 underestimating the reflectance of the \textit{Marbles}.
Only in the case of the \textit{Water bottle}, conventional methods have a relatively higher accuracy than CLIP-based methods, but lower accuracy than GPT methods.

Though CLIP (ViT-B/32) shows the highest accuracy, its mean error increases as the \textit{complexity} of the objects increase, from 6\% for the regular objects, to 10\% for the irregular objects and to 17\% for the transparent objects.
For regular and irregular non-transparent objects, GPT-4 seconds CLIP (ViT-B/32), while GPT-3.5 is second to last.
However, GPT-3.5 surpasses GPT-4 for transparent objects by about 2\%, the same margin that CLIP-based methods hold over GPT-3.5.

\begin{table*}[t]
\caption{Mean error of reflectance of infrared light}
\label{table:mae}
\centering
\begin{tabular}{clcccc}

\toprule

& Method  & Pre-training dataset & Known objects                         & Unseen objects                      &  \\

\midrule

\multirow{3}{*}{Conventional}
& VGG16     & ImageNet             & $ 0.021 \pm 0.025 $ & $ 0.249 \pm 0.145 $ &  \\
& ResNet101 & ImageNet             & $ \underline{0.008 \pm 0.008} $ & $ 0.178 \pm 0.143 $ &  \\
& ViT-B/32  & ImageNet             & $ 0.011 \pm 0.015 $ & $ 0.184 \pm 0.112 $ &  \\
\midrule
\multirow{2}{*}{VLM}
& CLIP (ResNet101) & WebImageText       & $ \underline{\mathbf{0.006 \pm 0.008}} $ & $ \underline{0.144 \pm 0.105} $ &  \\
& CLIP (ViT-B/32)  & WebImageText       & $ \underline{0.008 \pm 0.010} $ & $ \underline{\mathbf{0.118 \pm 0.094}} $ &  \\
\midrule
\multirow{2}{*}{LLM}
& GPT-3.5   & *                    & -                 & $ 0.199 \pm 0.255 $ &  \\
& GPT-4     & *                    & -                 & $ 0.147 \pm 0.150 $ \\

\bottomrule

\end{tabular}
\parbox{\textwidth}{
    \footnotesize
    \ \ *Not disclosed.
}
\end{table*}

\subsection{Object Grasping}
\label{sec:grasping}

\begin{figure*}
     \centering
     \includegraphics[width=0.8\textwidth]{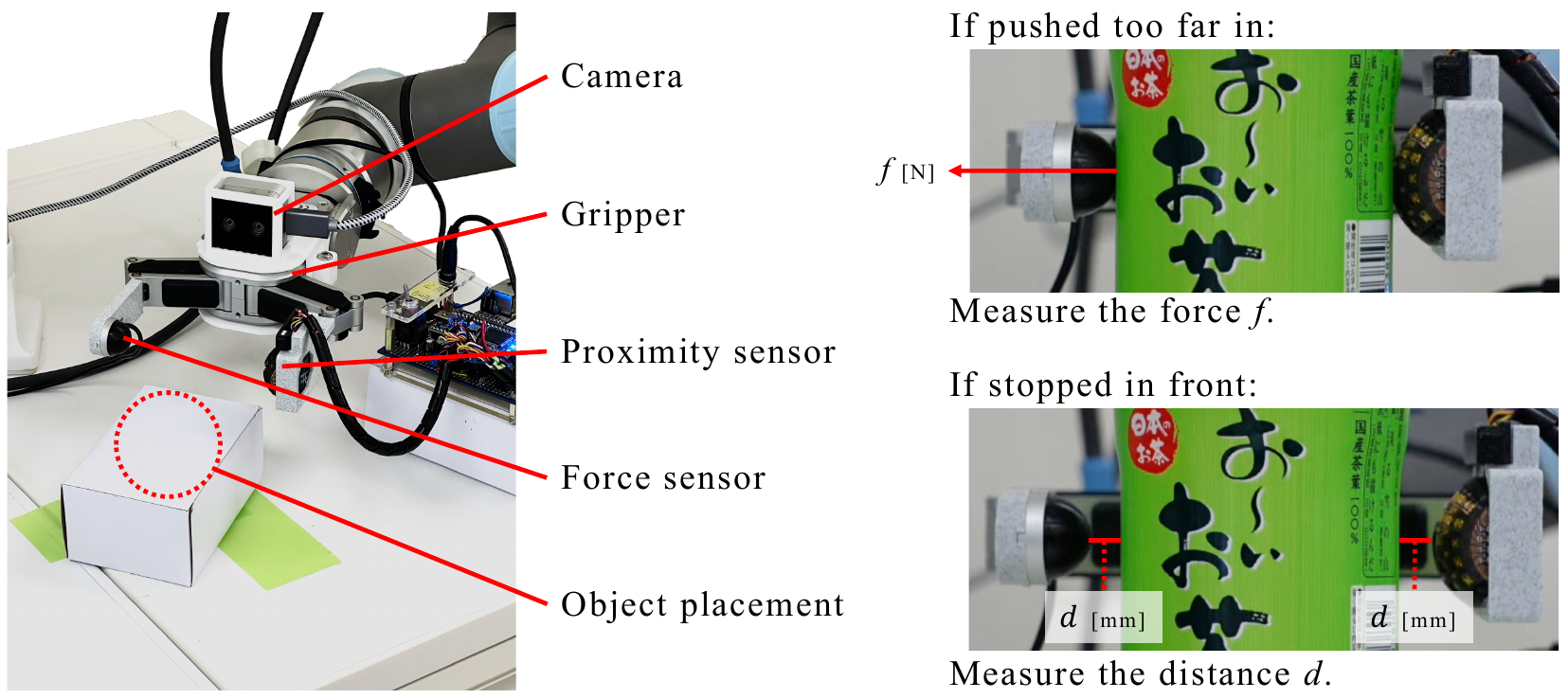}
     \caption{Experimental setup utilizing a two-finger gripper equipped with a proximity sensor in one fingertip and a force sensor in the opposite one.}
     \label{fig:grip_experiment_setting}
\end{figure*}

We perform an experiment using an actual gripper to validate the proposed method's effectiveness in the robotic grasping. 
As shown in Fig.~\ref{fig:grip_experiment_setting}, the two-finger gripper has a proximity sensor in one fingertip and a force sensor in the opposite fingertip.
We use the OnRobot RG2 gripper, which is a parallel two-finger gripper with a maximum opening width of 110 mm.
Moreover, we use the OptoForce OMD-20-SE-40N sensor, a 3-axis force sensor with a nominal capacity of $\pm20$ N, which we placed in the fingertip opposite to the proximity sensor.
3D-printed attachments were designed so that the distance from the tip of each sensor to the TCP of the gripper was the same.

In the experiments, the gripper closes the fingers so that the fingertips advance the distance estimated by the compared methods.
Should the reflectance estimation be incorrect, the robot may push into the object or stop prematurely. 
When the push exceeds the required distance, the force applied to the object is measured by the force sensor attached to the other fingertip. 
Conversely, if the robot stops too early, the discrepancy in the distance between the sensor and the object is manually measured. 
The measured force or distance error should be minimal if the reflectance is set appropriately.
The objects undergo five grasp repetitions using the test objects to observe the grasping result with each of the compared methods.
A force sensor measures the force the gripper applies, and a feeler gauge is used to measure the distance between the sensor and the object manually. 
Additionally, if an object is likely to deform significantly, any differences in distance are measured based on its original, undeformed state.

The experimental conditions have been categorized into two cases: one involves setting the reflectance to fixed values, while the other involves predicting it using a pre-/trained model. 
The latter method encompasses predictions utilizing  ResNet101, CLIP (ViT-B/32), and GPT-4.
We also include the actual measured reflectance serving as the ground truth. 
The inclusion of ground truth labels in the experiment aims to validate the effectiveness of the reflectance predictions.

\begin{figure*}
     \centering
     \includegraphics[width=0.8\textwidth]{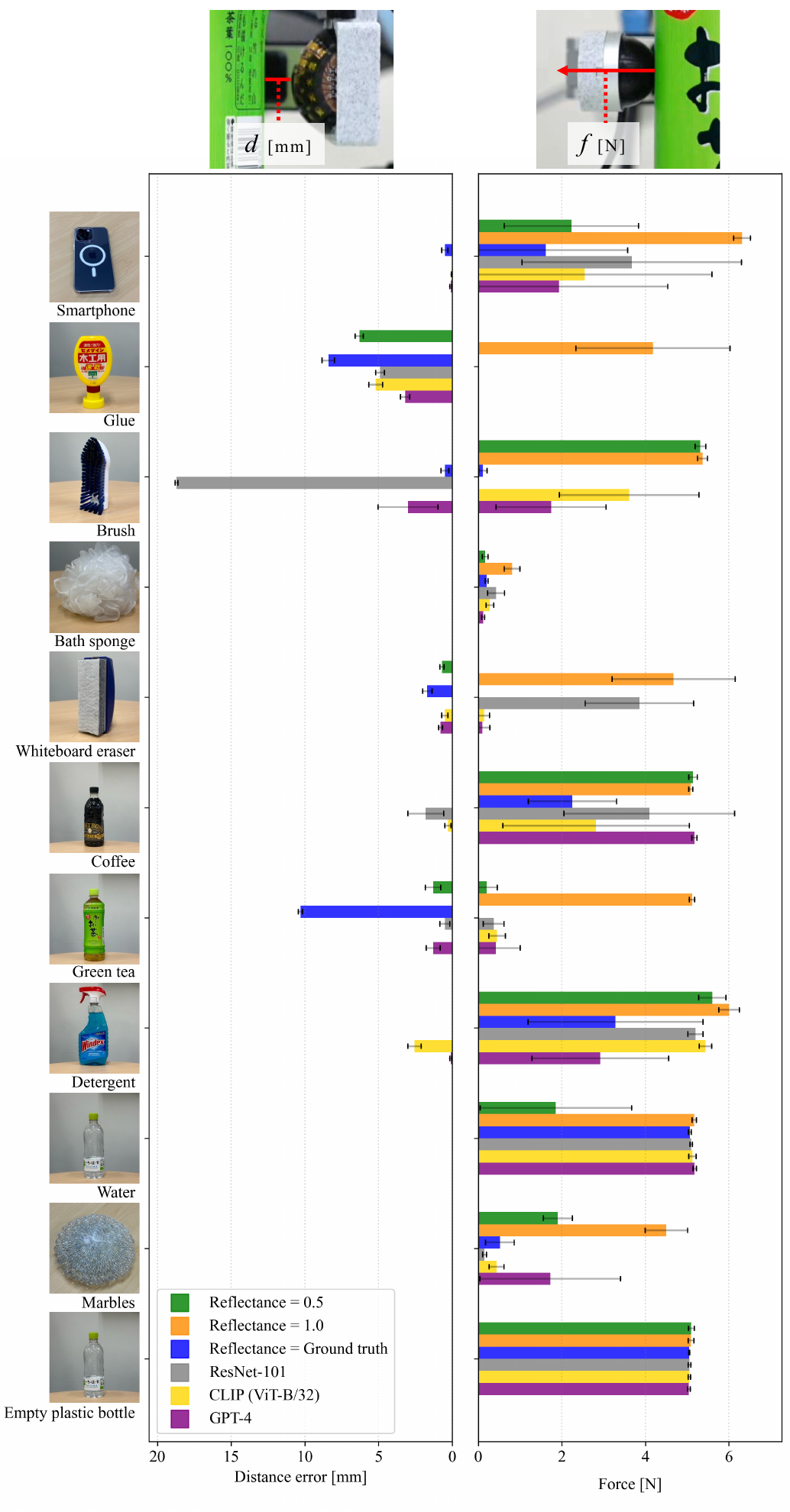}
     \caption{The effect of reflectance estimation on grasping an object.
        Left: distance error when distance is underestimated.
        Right: force exerted on the object when distance is overestimated.
        
     }
     \label{fig:grip_experiment_general}
\end{figure*}

Fig.~\ref{fig:grip_experiment_general} shows the results of the object grasping experiments.
Fig.~\ref{fig:grip_experiment_general} (left) shows the distance error corresponding to a distance underestimation, while Fig.~\ref{fig:grip_experiment_general} (right) shows the force exerted on the object corresponding to a distance overestimation.
Simply put, when there is a distance error, the object is not grasped, while if there is a force exerted on the object, the object is pressed on.

First, we focus on the results when the reflectance is set to a fixed value.
Using fixed reflectance values is used in conventional studies employing proximity sensing (e.g.,~\cite{hirai2018}).
In this paper, we set those values to 0.5 and 1.0, in respective experiments.
The results show that the effect of reflectance is a significant issue. 
In particular, with \textit{Glue}, setting the reflectance to 0.5 resulted in an average distance error of 6.30 mm, and with a reflectance of 1.0, the grasping force was 4.18 N. 
Additionally, when the reflectance is set to 1.0, the gripper presses into all objects.
This indicates that the distance is overestimated because the reflectance was set higher than it is.

Next, we focus on the comparison between ResNet, CLIP, and GPT-4.
CLIP and ResNet have been further trained on our dataset for reflectance prediction. 
However, GPT-4 has not been fine-tuned, and only a few correct labels have been provided through few-shot prompting. 
The results show that GPT-4 outperforms ResNet101 in seven of 11 objects, such as \textit{Smartphone}, \textit{Glue}, and \textit{Brush}, and outperforms CLIP(ViT-B/32) in six objects.
This result indicates that GPT-4, during its large-scale pre-training phase, has acquired practical knowledge about reflectance prediction comparable to or exceeding that of the additionally fine-tuned ResNet and CLIP.

Table~\ref{table:grip_experiment_general} shows a summary of the results of the grasping experiments for each method.
The closer the distance error and applied force are to zero, the more accurate the reflectance prediction, and the closer the gripper is to the grasping points.
There is a significant difference when the reflectance is fixed at 0.5 and 1.0, respectively, indicating that variations in reflectance affect object grasping.
With fixed reflectance at 0.5 (midpoint of the reflectance possible values), the distance error is similar to that of CLIP and GPT-4, but the force error is higher.
On the other hand, GPT-4 has more minor errors compared to ResNet-101. 
Additionally, the errors between CLIP and GPT-4 are not significantly different. 
It is noteworthy that GPT-4, which does not require additional training for this task, achieves a competitive performance using only text as input.

\begin{table}
\caption{The average effect of reflectance estimation on grasping an object}
\label{table:grip_experiment_general}
\centering
\begin{tabular}{lcc}
\toprule

Method & Distance error [mm]  & Force [N]   \\

\midrule

Fixed value (0.5)          & $  \underline{0.755 \pm 1.878} $ & $ 2.500 \pm 2.365 $  \\
Fixed value (1.0)          & - & $ 4.755 \pm 1.563 $  \\
\midrule
Ground truth            & $ 1.945 \pm 3.593 $ & $ \underline{\mathbf{1.644 \pm 2.124}} $  \\
\midrule
ResNet-101              & $ 2.355 \pm 5.485 $ & $ 2.537 \pm 2.451 $  \\
CLIP (ViT-B/32)         & $ \underline{\mathbf{0.749 \pm 1.686}} $ & $ 2.352 \pm 2.442 $  \\
GPT-4                   & $ 0.773 \pm 2.217 $ & $ \underline{2.211 \pm 2.297} $ \\

\bottomrule

\end{tabular}
\parbox{0.60\textwidth}{
    \footnotesize
    Because with Fixed value (1.0) the gripper exerted force onto the objects in all cases, there was no distance error.
}
\end{table}


\subsection{Discussion}
\label{sec:discussion}

In general, LLM/VLM methods outperformed the conventional approach in reflectance estimation; only GPT-3.5 was surpassed by the conventional ResNet101 and ViT-B/32.
While the superior performance of GPT-4 over GPT-3.5 shows that the simultaneous learning of language and vision is crucial, GPT-3.5 showed competitive results compared to supervised methods, even without using any image data.
This implies that the distributional semantics underlying LLMs trained only with text can be used to estimate reflectance, which is enhanced by the contribution of text and image training in the case of GPT-4.
(\textbf{RQ1})

The effect of distributional semantics can also be observed when comparing the CLIP methods with their conventional counterpart, e.g., CLIP (ResNet101) and ResNet101.
While being pre-trained on different datasets (ImageNet and WebImageText) could have an impact on the performance, one could assume that both datasets are sufficiently large and rich to make the (backbone) vision models equivalent.
Then, the only salient difference is the text-image training of CLIP, which resulted in higher generalization in reflectance estimation.
(\textbf{RQ2})

ViT-B/32 results tend to be the average of the reflectance values, while GPT-3.5 showed high variability.
On the other hand, CLIP methods displayed higher accuracy.
This shows that not using language information (image-only) leads to lower accuracy while relying only on language information for image-based tasks leads to more variable results.
When using both image and language information, a higher accuracy could be obtained.
(\textbf{RQ2})

Regarding the results for transparent objects, LLM/VLM methods show high accuracy even though the prompt contained no explicit information about their transparency.
This implies that the relationship between the objects and their physical properties is indeed embedded in the LLMs/VLMs, as distributional semantics suggests, and that such extrapolation capabilities increase their generalization capabilities.
(\textbf{RQ1} and \textbf{RQ2})

In regard to grasping, the experiments showed that VLMs have a tendency to overestimate distance, which made the gripper press on the object.
Interestingly, GPT-4 made the gripper press on the objects with the least force after the ground truth and led to distance errors within 5 mm in spite of using only text inputs.
This suggests that language embeds relatively accurate reflectance information that can be used in robotic grasping when using proximity sensors.
(\textbf{RQ1} and \textbf{RQ2})


\section{Conclusion}
\label{sec:conclusion}

We experimentally verified that LLMs such as GPT-3.5 and GPT-4 can estimate an object's reflectance using only text as input, and that VLMs such as CLIP can increase their generalization capabilities to achieve a higher accuracy in reflectance estimation from images.
This suggests that the tacit knowledge derived from distributional semantics that these models possess can be used for low-level cognition in robotics. 
This was clarified in the paper from two perspectives.
First, LLMs (GPT-3.5 and GPT-4) demonstrated a competitive performance when compared to existing methods through few-shot learning alone without using any image information.
Second, VLMs (CLIP-based methods) achieved higher accuracy than baseline methods that learn from images alone. 
This means that learning from language and the obtained distributional semantics helped to generalize the estimation of reflectance dependent on object classes from images.

The current work shows that language contributes to the generalization of reflectance estimation.
However, as language is an abstraction in itself, this could cause a bias induced by implicit categorization, which could result in ignoring subtle differences in the objects.
For example, \textit{bottle(s)} can have a virtually infinite number of differences but they may be inadvertently categorized as a single type of objects.
We plan to investigate to what extent this affects robotic grasping: while there may not be a significant effect on objects with monotonous surfaces, there may be significant effects on objects with particular variations.

Furthermore, the current work assumes a single reflectance value for the entire object but, in reality, objects can have multiple reflectance values across their surfaces.
The results presented in this paper could suggest that the knowledge embedded in LLMs contains the association that humans do between an object and a single reflectance, and that such association could be the limit of what LLMs can estimate.
We plan to investigate if a higher level of detail can be obtained using LLMs and VLMs to create dense 3D maps of the objects' reflectance.

Finally, as a similar approach can be followed for estimating other physical properties of objects, we plan to estimate the friction of the objects' surfaces, which is crucial to calculate the force needed to grasp an object without letting it slip or crushing it.


\section*{Acknowledgments}
\label{sec:acknowledgments}

This work was supported by the Japan Science and Technology Agency~(JST), Moonshot Research \& Development Program, Grant Number JPMJMS2011.


\section*{Disclosure Statement}
\label{sec:disclosure}

The authors report that there are no competing interests to declare.


\section*{Authors}
\label{sec:biography}

\begin{description}
    \item[Masashi Osada] received the B.Eng. degree in Information Science from Kanazawa Institute of Technology, Ishikawa, Japan, in 2023.
    He is currently pursuing a Master's degree at Ritsumeikan University, Japan.
    His research interests include machine learning, particularly deep learning, and its real-world applications.
    \\
    \item[Gustavo Alfonso Garcia Ricardez] received his M.Eng. and Ph.D. degrees from the Nara Institute of Science and Technology, Japan, in 2013 and 2016, respectively.
    He is currently an Associate Professor at Ritsumeikan University, and a Research Advisor for Robotics Competitions at the Robotics Hub of Panasonic Corporation, where he leads numerous research projects and teams in international robotics competitions.
    His research interests include human-safe, efficient robot control, human-robot interaction, manipulation, and task planning.
    \\
    \item[Yosuke Suzuki] received the B.Eng., M.Eng., and Ph.D. degrees in engineering from the Tokyo Institute of Technology, Tokyo, Japan, in 2005, 2007, and 2010, respectively.
    He is currently an Associate Professor with the Faculty of Mechanical Engineering, Institute of Science and Engineering, Kanazawa University, Kanazawa, Japan.
    His research interests include tactile and proximity sensors, robotic grasping, and distributed autonomous systems.
    \\
    \item[Tadahiro Taniguchi] received his M.Eng. and Ph.D. degrees from Kyoto University, Japan, in 2003 and 2006, respectively.
    From 2005 to 2008, he was a Japan Society for the Promotion of Science Research Fellow in the same university.
    From 2008 to 2010, he was an Assistant Professor at the Department of Human and Computer Intelligence, Ritsumeikan University, Japan.
    From 2010 to 2017, he was an Associate Professor in the same department.
    From 2015 to 2016, he was a Visiting Associate Professor at the Department of Electrical and Electronic Engineering, Imperial College London, UK.
    Since 2017, he is a Professor at the Department of Information and Engineering, Ritsumeikan University, and a Visiting General Chief Scientist at the Technology Division of Panasonic Corporation.
    He has been engaging in research on machine learning, emergent systems, intelligent vehicle, and symbol emergence in robotics.
\end{description}


\bibliographystyle{templates/tADR}
\bibliography{tADR}


\newpage


\appendix

\section{Proximity Sensor Output}
\label{sec:appendix-sensor-outputs}

\begin{figure}[h]
    \begin{center}
        \subfigure[
            Regular objects.
        ]{
            \includegraphics[height=5.0cm]{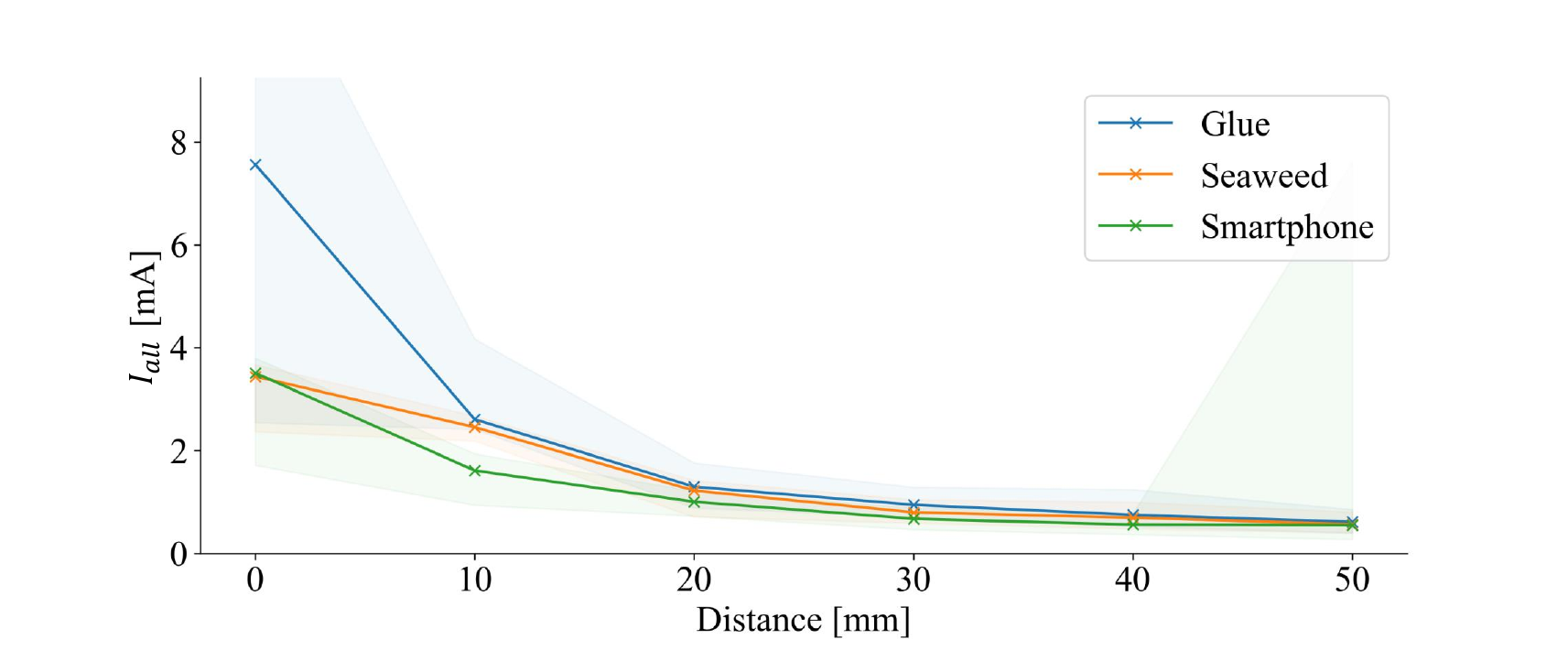}
        }
        \subfigure[
            Irregular objects.
        ]{
            \includegraphics[height=5.0cm]{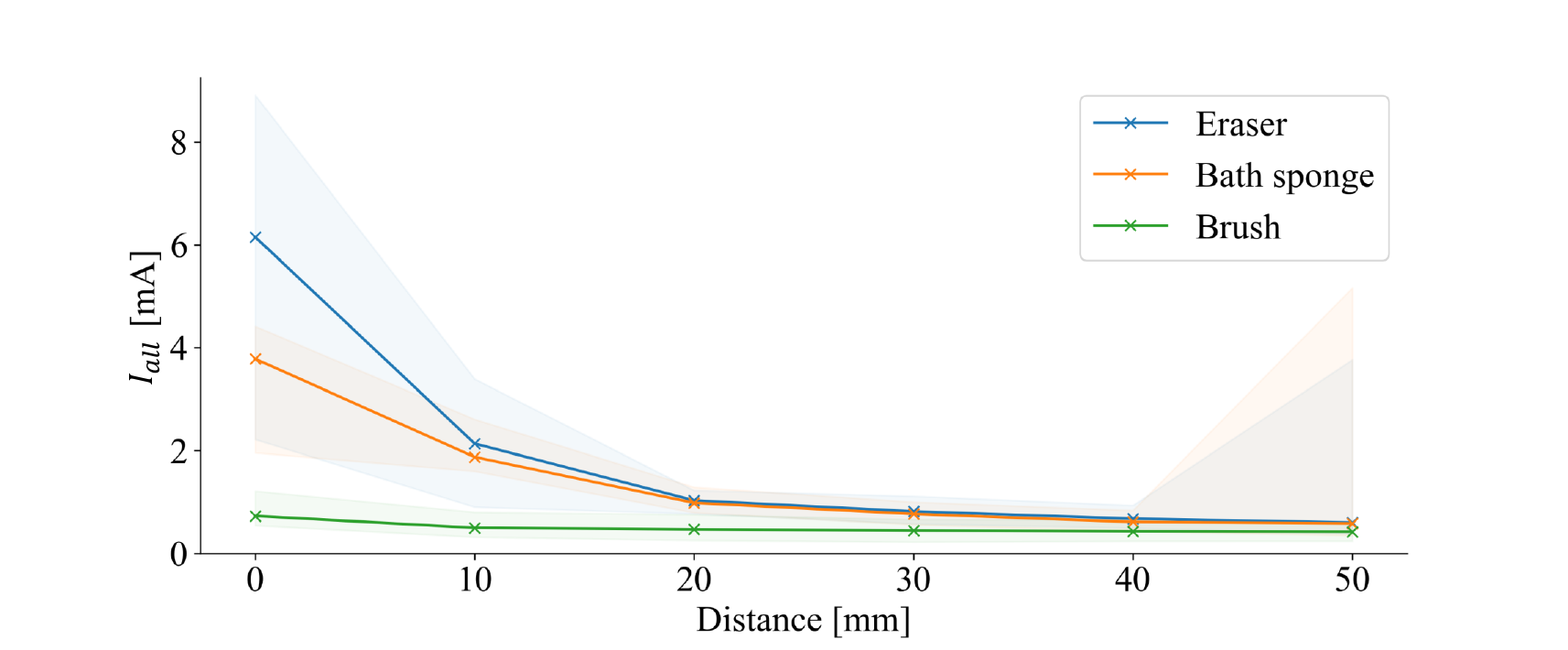}
        }
        \subfigure[
            Transparent objects.
        ]{
            \includegraphics[height=5.0cm]{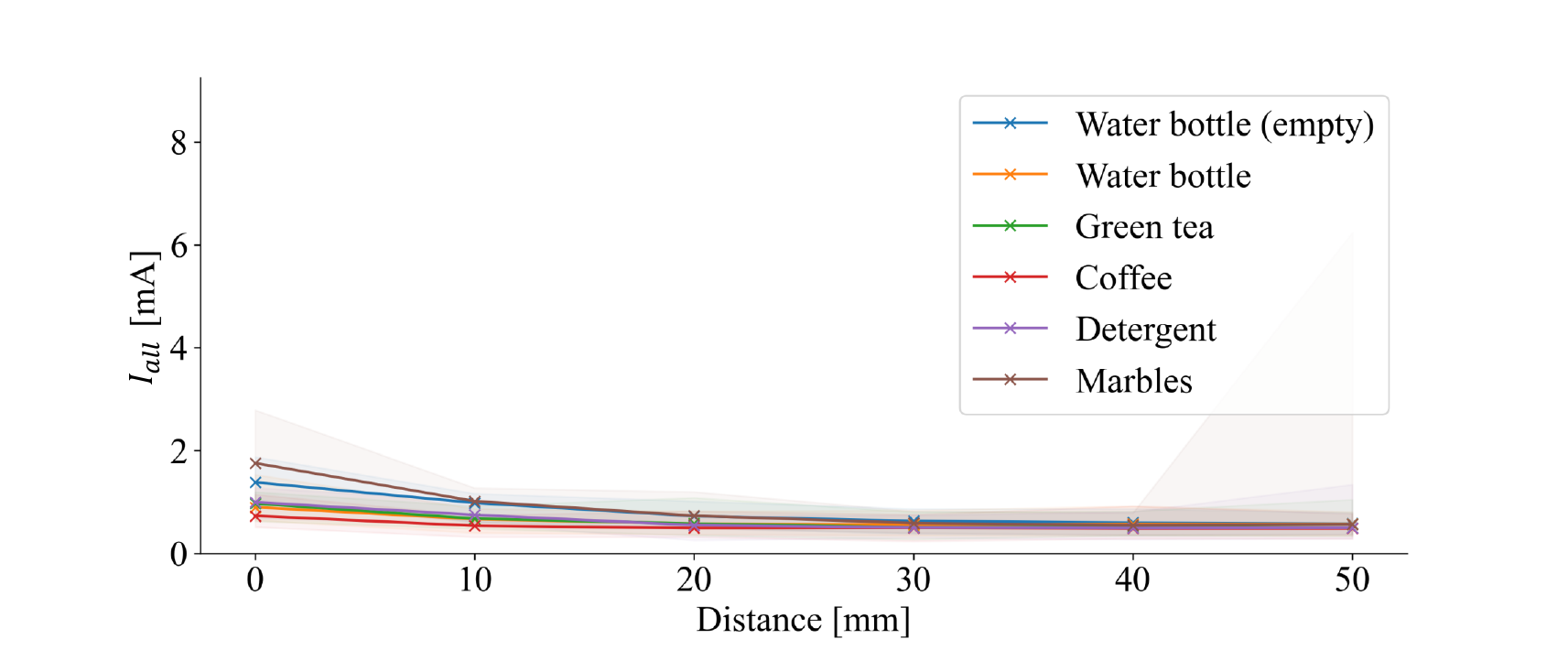}
        }
        \caption{
           Results of $I_{all}$ for the calculation of the reflectance ground truth.
        }
        \label{fig:Iall-results}
    \end{center}
\end{figure}

Fig.~\ref{fig:Iall-results} shows the results of the measurements obtained with the proximity sensor.
The vertical axis corresponds to $I_{all}$, which is the sum of the current from all the phototransistors in the proximity sensors.

During training, we calculated and used the reflectance from the average value of the proximity sensor output from 200 samples per object.
This is because we aim to estimate the reflectance as a single real number, both from the trainable models and from the pre-trained models such as GPT-4.
However, it could be possible to use distributional regression networks (e.g.,~\cite{distributional-regression}) to estimate reflection accounting for the variance of the sensor output.

\newpage
\section{Prompt to Estimate Reflectance with GPT-3.5 and GPT-4}
\label{sec:appendix-prompt}

The pseudocode of the prompts and an example of the replies are shown below.
To improve readability, the prompt specifies that the output should be formatted as a table in markdown.

\begin{mdframed}
\begin{verbatim}
You are an expert in material properties. Your task is to determine the 
infrared reflectance of a given object. Please provide a single, reasona-
ble value from the estimated range of infrared reflectance for each ob-
ject. If your prediction is not conclusive, please output an approximate
value as a guess. Note that the reflectivity of plastic bottles varies 
depending on their contents and labels.

Here are some example objects and their reflectance for reference:

salad dressing bottles [plastic, orange] : 0.326
salad dressing bottles [plastic, brown] : 0.327
salad dressing bottles [plastic, grean] : 0.399
salad dressing bottles [plastic, yellow] : 0.403
...

---

User: get_reflectance(["energy drink", "yogurt"])
You:
|item|result|
|:--|:--|
|Name|energy drink|
|Reason|Aluminum is a metal, but its surface is printed with resin ink, 
which reflects light less than cardboard or mirrors. Since the reflectance 
of the chips tube is 0.684, the reflectance of an energy drink (aluminum 
can) is estimated to be 0.422.|
|**Response**|range: 0.41 - 0.46, prediction: **0.422**|

|item|result|
|:--|:--|
|Name|yogurt|
|Reason|Yogurt is a food product, but this is usually packaged in a plastic 
container. The yogurt container is considered less reflective than a 
similarly labeled plastic bottle. So, the reflectance of yogurt is greater 
than 0.616. On the other hand, it is white in color and will reflect 
relatively well. Therefore, we can expect it to be 0.75.|
|**Response**|range: 0.68 - 0.79, prediction: **0.762**|
\end{verbatim}
\end{mdframed}


\newpage
\section{Reflectance Estimation for Known Objects}
\label{sec:appendix-reflectance-known}

\begin{figure}[h]
     \centering
         \includegraphics[width=\textwidth]{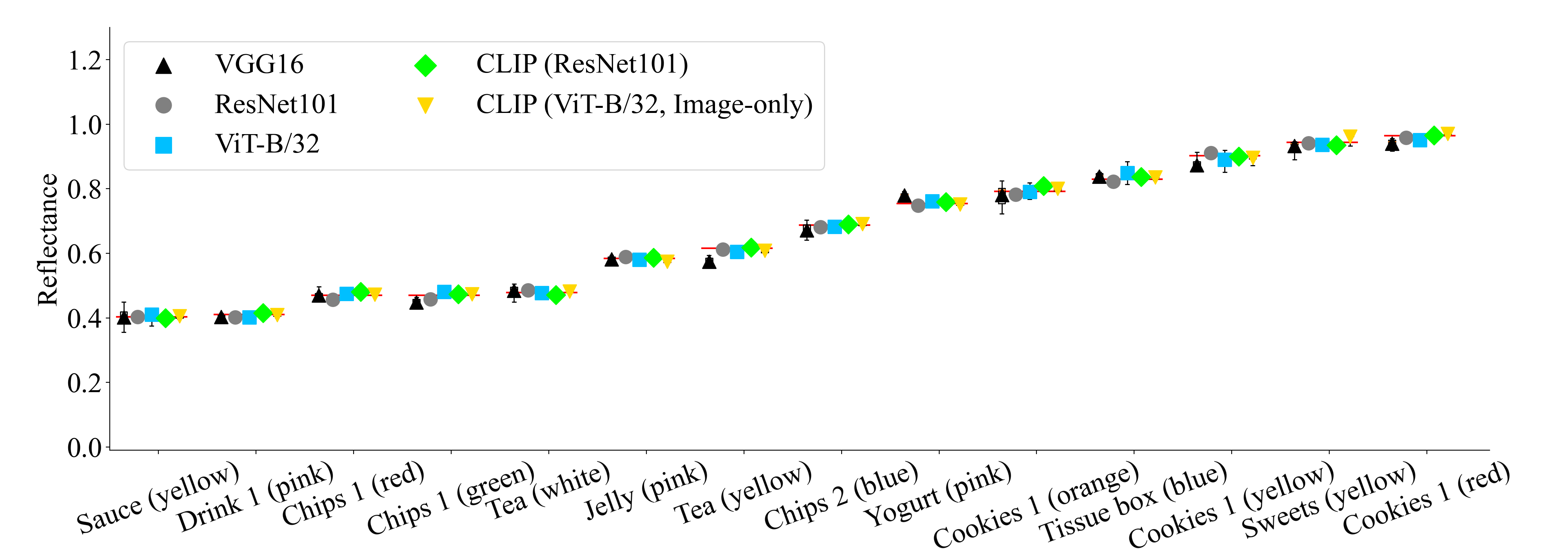}
         \caption{Estimated reflectance for known objects (training set).}
         \label{fig:visual_compare_known}
\end{figure}

Fig.~\ref{fig:visual_compare_known} shows the estimated reflectance for known objects (training set).
In all cases, the image-only VGG16, ResNet101 and ViT-B/32 as well as the CLIP-based methods show a high accuracy on the training dataset.
The results of GPT methods are omitted because the reference values for the training set were provided in the prompt, causing GPT methods to return exactly the same values.

\newpage
\section{Detailed Results of Object Grasping Experiments}

\begin{figure}[h]
     \centering
         \includegraphics[width=\textwidth]{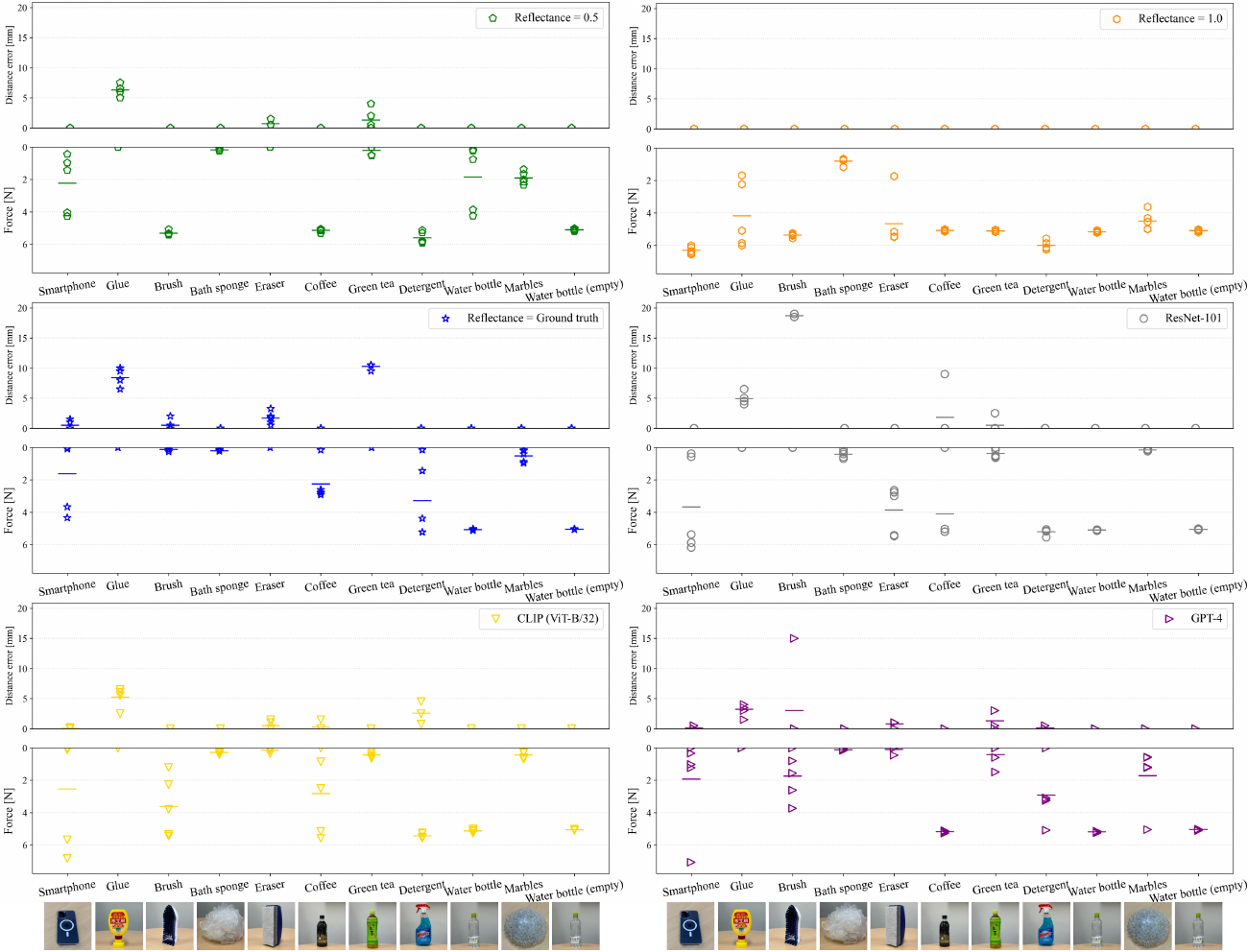}
         \caption{Effect of reflectance estimation on object grasping for the compared methods}
         \label{fig:grip_experiment_by_method}
\end{figure}

\begin{figure}[h]
     \centering
         \includegraphics[width=0.9\textwidth]{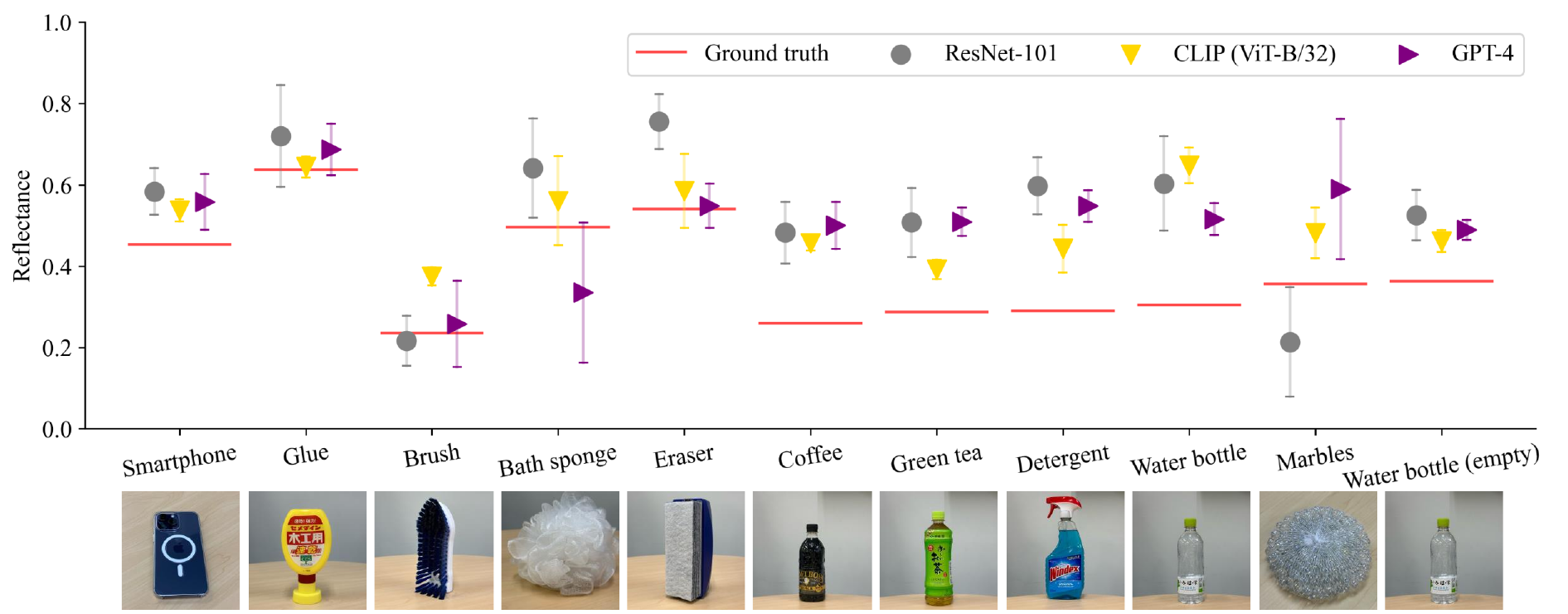}
         \caption{Estimated reflectance for objects used in the experiment}
         \label{fig:grip_experiment_prediction}
\end{figure}

Fig.~\ref{fig:grip_experiment_by_method} details the results shown in Fig.~\ref{fig:grip_experiment_general} (main text). 
Each point on the graph represents the force or distance recorded in each experiment, and the line segments indicate their mean values. 
Fig.~\ref{fig:grip_experiment_prediction} depicts the predicted reflectance for the objects used in the experiments. 
Additionally, the ground truth is known since the test set is used for the objects.

\begin{table}
\caption{Average distance error [mm] per object in the grasping experiments}
\label{table:grip_experiment_detail_distance}
\centering
\footnotesize
\begin{tabular}{lcccccc}
\toprule

Object & Fixed (0.5) & Fixed (1.0) & Ground truth & ResNet-101 & CLIP(ViT) & GPT-4\\

\midrule
Smartphone           &                      - &                      - &    $ 0.500 \pm 0.632 $ &                      - &\underline{$\mathbf{0.040 \pm 0.080}$} &    $ \underline{0.100 \pm 0.200} $ \\
Glue                 &    $ 6.300 \pm 0.812 $ &                      - &    $ 8.400 \pm 1.241 $ &    $ \underline{4.900 \pm 0.860} $ &    $ 5.200 \pm 1.400 $ &\underline{$\mathbf{3.200 \pm 0.927}$}\\
Brush                &                      - &                      - &\underline{$\mathbf{0.500 \pm 0.775}$} &   $ 18.700 \pm 0.245 $ &                      - &    $ \underline{3.000 \pm 6.000} $ \\
Bath sponge          &                      - &                      - &                      - &                      - &                      - &                      - \\
Eraser               &    $ \underline{0.700 \pm 0.400} $ &                      - &    $ 1.700 \pm 0.941 $ &                      - &\underline{$\mathbf{0.500 \pm 0.632}$}&    $ 0.800 \pm 0.400 $ \\
Coffee               &                      - &                      - &                      - &    $ \underline{1.800 \pm 3.600} $ &\underline{$\mathbf{0.300 \pm 0.600}$}&                      - \\
Green tea            &    $ 1.300 \pm 1.536 $ &                      - &   $ 10.300 \pm 0.400 $ &\underline{$\mathbf{0.500 \pm 1.000}$}&                      - &    $ \underline{1.300 \pm 1.400} $ \\
Detergent            &                      - &                      - &                      - &                      - &    $ \underline{2.562 \pm 1.327} $ &\underline{$\mathbf{0.100 \pm 0.200}$}\\
Water bottle         &                      - &                      - &                      - &                      - &                      - &                      - \\
Marbles              &                      - &                      - &                      - &                      - &                      - &                      - \\
Water bottle (empty) &                      - &                      - &                      - &                      - &                      - &                      - \\
\bottomrule

\end{tabular}
\parbox{\textwidth}{
    \footnotesize
}
\end{table}

\begin{table}
\caption{Average force [N] applied per object in the grasping experiments}
\label{table:grip_experiment_detail_force}
\centering
\footnotesize
\begin{tabular}{lcccccc}
\toprule

Object & Fixed (0.5) & Fixed (1.0) & Ground truth & ResNet-101 & CLIP(ViT) & GPT-4\\

\midrule
Smartphone           &    $ 2.226 \pm 1.607 $ &    $ 6.314 \pm 0.204 $ &\underline{$\mathbf{1.615 \pm 1.960}$}&    $ 3.674 \pm 2.628 $ &    $ 2.543 \pm 3.052 $ &    $ \underline{1.929 \pm 2.604} $ \\
Glue                 &                      - &    $ \underline{\mathbf{4.182 \pm 1.846}} $ &                      - &                      - &                      - &                      - \\
Brush                &    $ 5.314 \pm 0.125 $ &    $ 5.367 \pm 0.119 $ &\underline{$\mathbf{0.110 \pm 0.099}$}&                      - &    $ 3.612 \pm 1.671 $ &    $ \underline{1.741 \pm 1.316} $ \\
Bath sponge          &    $ \underline{0.164 \pm 0.067} $ &    $ 0.807 \pm 0.185 $ &    $ 0.199 \pm 0.033 $ &    $ 0.422 \pm 0.202 $ &    $ 0.276 \pm 0.091 $ &\underline{$\mathbf{0.113 \pm 0.035}$}\\
Eraser               &                      - &    $ 4.673 \pm 1.472 $ &                      - &    $ 3.857 \pm 1.300 $ &    $ \underline{0.134 \pm 0.131} $ &\underline{$\mathbf{0.090 \pm 0.181}$}\\
Coffee               &    $ 5.140 \pm 0.099 $ &    $ 5.082 \pm 0.048 $ &\underline{$\mathbf{2.251 \pm 1.056}$}&    $ 4.090 \pm 2.047 $ &    $ \underline{2.818 \pm 2.235} $ &    $ 5.173 \pm 0.059 $ \\
Green tea            &\underline{$\mathbf{0.203 \pm 0.249}$}&    $ 5.115 \pm 0.059 $ &                      - &    $ \underline{0.366 \pm 0.249} $ &    $ 0.448 \pm 0.198 $ &    $ 0.416 \pm 0.583 $ \\
Detergent            &    $ 5.600 \pm 0.326 $ &    $ 6.005 \pm 0.247 $ &    $ \underline{3.281 \pm 2.095} $ &    $ 5.198 \pm 0.180 $ &    $ 5.436 \pm 0.146 $ &\underline{$\mathbf{2.918 \pm 1.636}$}\\
Water bottle         &\underline{$\mathbf{1.855 \pm 1.811}$}&    $ 5.165 \pm 0.051 $ &    $ \underline{5.066 \pm 0.032} $ &    $ 5.095 \pm 0.028 $ &    $ 5.122 \pm 0.086 $ &    $ 5.178 \pm 0.038 $ \\
Marbles              &    $ 1.902 \pm 0.348 $ &    $ 4.500 \pm 0.508 $ &    $ 0.518 \pm 0.342 $ &\underline{$\mathbf{0.147 \pm 0.047}$}&    $ \underline{0.437 \pm 0.176} $ &    $ 1.721 \pm 1.684 $ \\
Water bottle (empty) &    $ 5.099 \pm 0.069 $ &    $ 5.090 \pm 0.069 $ &    $ \underline{5.047 \pm 0.013} $ &    $ 5.056 \pm 0.033 $ &    $ 5.051 \pm 0.028 $ &\underline{$\mathbf{5.039 \pm 0.032}$}\\
\bottomrule

\end{tabular}
\parbox{\textwidth}{
    \footnotesize
}
\end{table}

Tables~\ref{table:grip_experiment_detail_distance} and~\ref{table:grip_experiment_detail_force} show the mean and standard deviation of the distance error and exerted force for each object, respectively.
In Table~\ref{table:grip_experiment_detail_distance}, the hyphen indicates that in all trials there was force exerted on the tested object, hence the absence of distance data.
Similarly, in Table~\ref{table:grip_experiment_detail_force}, the hyphen indicates that in all trials there was a distance error, hence the absence of force data.
When values are shown in both tables for the same object and method, it means that some trials resulted in distance errors while others in exerted force.
Tables~\ref{table:grip_experiment_distance_significance} and~\ref{table:grip_experiment_force_significance} show the $p$-values of each grasping method in terms of distance error and exerted force, respectively.

\begin{table}[ht]
\centering
\small
\caption{$p$-values for each grasping method (distance error)}
\label{table:grip_experiment_distance_significance}
\begin{tabular}{lccccc}
\toprule
                & Fixed (0.5) & Fixed (1.0) & Ground truth & ResNet-101   & CLIP(ViT-B/32) \\
\midrule
Fixed (1.0)     & -           &             &              &              &               \\
Ground truth    & 0.280       & -           &              &              &               \\
ResNet-101      & *           & -           & 0.322        &              &               \\
CLIP(ViT-B/32)  & 0.900       & -           & 0.532        & *            &               \\
GPT-4           & 0.900       & -           & 0.345        & **           & 0.892         \\
\bottomrule
\end{tabular}
\parbox{0.93\textwidth}{
    \footnotesize
    *) $ p < 0.05 $ , **) $ p < 0.01 $ \\
    Because with Fixed value (1.0) the gripper exerted force onto the objects in all cases, there was no distance error.
}
\end{table}

\begin{table}[ht]
\caption{$p$-values for each grasping method (force)}
\label{table:grip_experiment_force_significance}
\centering
\small
\begin{tabular}{lccccc}
\toprule
                & Fixed (0.5) & Fixed (1.0) & Ground truth & ResNet-101   & CLIP(ViT-B/32) \\
\midrule
Fixed (1.0)  & **             &                &              &              &                \\
Ground truth    & 0.230          & *              &              &              &                \\
ResNet-101      & 0.900          & **             & 0.286        &              &                \\
CLIP(ViT-B/32)  & 0.900          & *              & 0.743        & 0.900        &                \\
GPT-4           & 0.900          & *              & 0.643        & 0.900        & 0.900          \\
\bottomrule
\end{tabular}
\parbox{0.93\textwidth}{
    \footnotesize
    *) $ p < 0.05 $ , **) $ p < 0.01 $
}
\end{table}

\newpage

\section{Mean Error of Reflectance Estimation with Image, Text and Both}
\label{sec:appendix-estimation-including-text}

\begin{table}[th]
\caption{Mean error of reflectance of infrared light}
\label{table:mae_including_text}
\centering
\begin{tabular}{llllc}
\toprule
Backbone       & Pre-training dataset & Input modal    & Combined by     & Unseen objects    \\
\midrule
ViT-B/32       & ImageNet             & Image-only     & -             & $ 0.184 \pm 0.108 $ \\
ViT-B/32 & WebImageText         & Image-only     & -             & $ 0.118 \pm 0.086 $ \\
ViT-B/32 & WebImageText         & Image and text & Addition      & $ \underline{0.082 \pm 0.086} $ \\
ViT-B/32 & WebImageText         & Image and text & Concatenation & $ \underline{\mathbf{0.079 \pm 0.084 }}$ \\
Transformer    & WebImageText         & Text-only      & -             & $ 0.098 \pm 0.087 $ \\
\bottomrule

\end{tabular}
\end{table}

In this paper, reflectance was predicted by inputting images into a VLM.
In this section, we attempt to input images and text into the VLM. 
The text input is a simple sentence that includes the name and color of the object.

Combining the information from the image and text encoders is necessary when making predictions. 
Therefore, in this experiment, we conducted training in two cases: one where the features extracted by each encoder were simply added together and another where the features were concatenated.
The outputs of both CLIP encoders are 512-dimensional vectors, so they remain 512-dimensional. However, in concatenation, they become 1024-dimensional. 
Passing this through a fully connected layer is mapped to the reflectance format.

\begin{figure}[h]
     \centering
         \includegraphics[width=\textwidth]{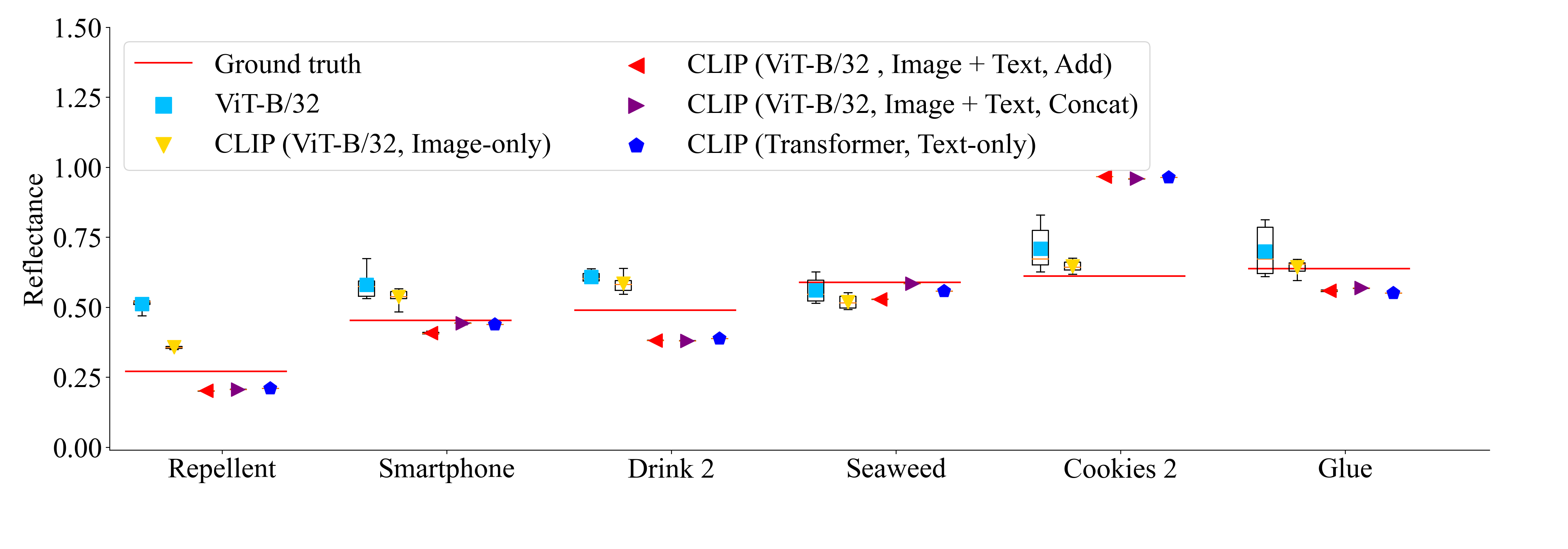}
         \caption{Estimated reflectance for regular objects in test set}
         \label{fig:detailed_error_include_text}
\end{figure}

The highest accuracy was obtained when both image and text were used and the information of the encoders was concatenated, as shown in Table~\ref{table:mae_including_text}.
Fig.~\ref{fig:detailed_error_include_text} shows detailed prediction results for regular objects. 
From this, we can see that models including language input are significantly less accurate for particular objects, such as \textit{Cookies 2}. 
These predictions closely resemble the reflectance of \textit{Cookies 1} included in the training data.
In the case of image input, these can be clearly distinguished. 
However, both are referred to as \textit{cookies} and \textit{cardboard boxes}, sharing similar information in language. 
Therefore, the training data heavily influences the predictions, leading to errors.

Understanding objects' material and appearance characteristics is important for estimating reflectance, and image input was very effective. 
Since VLM extracts standard features from images and language, it was shown that high-accuracy predictions are possible even from language input alone. 
On the other hand, it was also found that captions in language can deviate from actual, real-world images.

\newpage

\section{Comparison of Processing Speed and Number of Parameters}
\label{sec:appendix-processing-speed}

We measured the time it takes to estimate the reflectance value of a single object based on its image or name, and summarize it in Table~\ref{table:processing_speed}.
We also include in the table the number of parameters of the compared models, to the extent of what is disclosed.
This experiment used deep learning models, such as CLIP, running in a local environment with an RTX3070 and 8GB of RAM. 
Additionally, GPT was measured via ChatGPT.

\begin{table}[h]
\caption{Processing speed per object and parameter comparison of various backbones}
\label{table:processing_speed}
\centering
\begin{tabular}{lcccc}
\hline
Backbone  && Processing speed [s]     && Parameters     \\
\hline
VGG16      && $ 0.041 $ && $ 14,733,185 $  \\
ResNet101* && $ 0.040 $ && $ 23,661,505 $  \\
ViT-B/32*  && $ 0.037 $ && $ 88,073,537 $  \\
\hline
GPT-3.5    && $ 3.132 $ && ** \\
GPT-4      && $ 5.901 $ && ** \\
\hline
\end{tabular}
\parbox{10cm}{
    \footnotesize
    * \ The speed of CLIP remains the same because its backbone is consistent. \\
    \ \ ** Not disclosed.
}
\end{table}

\end{document}